# Gabor Convolutional Networks

Shangzhen Luan, Chen Chen , Baochang Zhang
Jungong Han and Jianzhuang Liu

***Abstract*—In steerable filters, a filter of arbitrary orientation can be generated by a linear combination of a set of "basis filters." Steerable properties dominate the design of the traditional filters, e.g., Gabor filters and endow features the capability of handling spatial transformations. However, such properties have not yet been well explored in the deep convolutional neural networks (DCNNs). In this paper, we develop a new deep model, namely, Gabor convolutional networks (GCNs or Gabor CNNs), with Gabor filters incorporated into DCNNs such that the robustness of learned features against the orientation and scale changes can be reinforced. By manipulating the basic element of DCNNs, i.e., the convolution operator, based on Gabor filters, GCNs can be easily implemented and are readily compatible with any popular deep learning architecture. We carry out extensive experiments to demonstrate the promising performance of our GCNs framework, and the results show its superiority in recognizing objects, especially when the scale and rotation changes take place frequently. Moreover, the proposed GCNs have much fewer network parameters to be learned and can effectively reduce the training complexity of the network, leading to a more compact deep learning model while still maintaining a high feature representation capacity. The source code can be found at https://github.com/bczhangbczhang.**

## I. INTRODUCTION

ANISOTROPIC filtering techniques have been widely used to extract robust image representation. In particular, Gabor filter, which is based on a sinusoidal plane wave, has been explored extensively in various applications such as face recognition, texture classification, since it can characterize the spatial frequency structure in images while preserving information of spatial relations. This enables us to extract orientation-dependent frequency contents of patterns. Moreover, steerable and scalable kernels like Gabor functions can be easily specified and are computationally efficient. Recently, deep convolutional neural networks (DCNNs) based on convolution filters have attracted significant attention in computer vision due to the amazing capability of learning powerful feature representations from raw image pixels. Unlike hand-crafted filters with no learning process involved, DCNNs-based feature extraction is a data-driven technique that can *learn* robust feature representations from data directly. However, it comes with the cost of expensive training and complex model parameters, e.g., a typical CNN has millions of parameters to be optimized and performs more than 1B high precision operations to classify one image. Additionally, DCNNs' limited capability of modeling geometric transformations mainly comes from extensive data augmentation, large models, and hand-crafted modules (*e.g.* max-pooling [1] for small translation-invariance), therefore they normally fail to handle large and unknown object transformations if the training data are not enough, which is very likely in many real-world applications. And one reason originates from the way of filter designing [1], [2] with fixed receptive field size at every location in one convolution layer. There is a lack of internal mechanism to deal with the geometric transformations.

Fortunately, enhancing model capacity to transformations via filter redesigning has been acknowledged by researchers and some attempts have been made in recent years. Existing approaches basically follow two directions: deformable filter and rotating filer. In [3], a deformable convolution filter was introduced to enhance DCNNs' capacity of modeling geometric transformations by allowing free form deformation of the sampling grid with offsets learned from the preceding feature maps. However, the deformable filtering is complicated, because it is always associated with the Region of Interest (RoI) pooling technique originally designed for object detection [4]. Besides, the deformable convolution and deformable ROI pooling modules add extra amount of parameters and computation for the offset learning, which further increase the model complexity. In [2], Actively Rotating Filters (ARFs) were proposed to improve the generalization ability of DCNNs against rotation. The ARFs actively rotate during convolution so as to produce feature maps with location and orientation explicitly encoded. However, such a filter rotation method may be only suitable for small and simple

Manuscript received September 14, 2017; revised April 4, 2018; accepted April 29, 2018. Date of publication May 10, 2018; date of current version June 1, 2018. This work was supported in part by the Natural Science Foundation of China under Contract 61672079 and Contract 61473086, in part by the Shenzhen Peacock Plan under Grant KQTD2016112515134654, in part by the Open Projects Program of National Laboratory of Pattern Recognition, and in part by the National Basic Research Program of China under Grant 2015CB352501. The associate editor coordinating the review of this manuscript and approving it for publication was Prof. Xudong Jiang.
*(Corresponding authors: Baochang Zhang; Jungong Han.)*

S. Luan is with the School of Automation Science and Electrical Engineering, Beihang University, Beijing 100191, China (e-mail: luanshangzhen@buaa.edu.cn).

C. Chen is with the Department of Electrical and Computer Engineering, The University of North Carolina at Charlotte, Charlotte, NC 28223 USA (e-mail: chenchen870713@gmail.com).

B. Zhang is with the School of Automation Science and Electrical Engineering, Beihang University, Beijing 100191, China, and also with the Shenzhen Academy of Aerospace Technology, Shenzhen 518057, China (e-mail: bczhang@buaa.edu.cn).

J. Han is with the School of Computing and Communications, Lancaster University, Lancaster LA1 4YW, U.K. (e-mail: jungonghan77@gmail.com).

J. Liu is with the Noah's Ark Lab, Huawei Technologies Co. Ltd., Shenzhen 518129, China (e-mail: liu.jianzhuang@huawei.com).

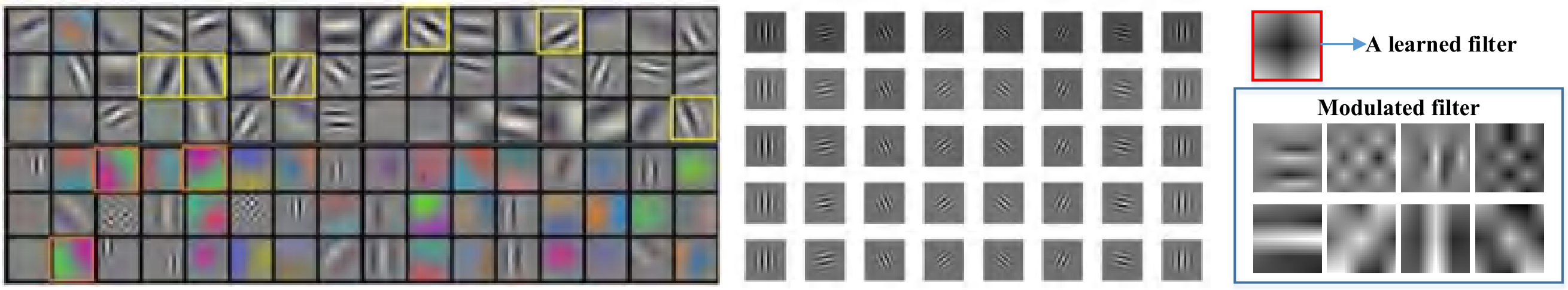

Fig. 1. Left illustrates AlexNet filters. Middle shows Gabor filters. Right presents the convolution filters modulated by Gabor filters. Filters are often redundantly learned in CNN, and some of which are similar to Gabor filters (see the highlighted ones with yellow boxes). Based on this observation, we are motivated to manipulate the learned convolution filters using Gabor filters, in order to achieve a compressed deep model with reduced number of filter parameters. In the right column, a convolution filter is modulated by Gabor filters via Eq. 2 to enhance the orientation property.

filters, *i.e.* $1 \times 1$ and $3 \times 3$ filters. While a general modulation method based on the Fourier transform was claimed, it was not implemented in [2], which is probably due to its computational complexity. Furthermore, 3D filters [5] are hardly modified by deformable filters or ARFs. In [6], by combining low level filters (Gaussian derivatives up to the 4-th order) with learned weight coefficients, the regularization over the filter function space is shown to improve the generalization ability but only when the set of training data is small.

### A. Motivation

To help us understand what CNN learns from the image data, we visualize the convolutional filters in Fig. 1 using AlexNet [7] trained on ImageNet[1] as an example. In this figure, we show the convolution filters from the first convolution layer. As can be seen, the learned filters are redundant in the shallow layers, and many filters are similar to Gabor filters (see the filters highlighted by yellow boxes in the left column of Fig. 1). For example, those filters with different orientations are able to capture edges in different directions. It is known that the steerable properties of Gabor filters are widely adopted in the traditional filter design due to their enhanced capability of scale and orientation decomposition of signals, which is unfortunately neglected in most of the prevailing convolutional filters in DCNNs. A few works have related Gabor filters to DCNNs. However, they do not explicitly integrate Gabor filters into the convolution filters. Specifically, Hu *et al.* [8] simply employ Gabor filters to generate Gabor features and use them as input to a CNN, and Sarwar *et al.* [9] only adopt the Gabor filters in the first or second convolution layers with the aim to reduce the training complexity of CNNs. Can we learn a small set of filters, which are then manipulated to create more in a similar way as the design of Gabor filters? If it is possible, one of the obvious advantages lies in that only a small set of filters are required to learn for DCNNs, leading to a more compact yet enhanced deep model.

### B. Contributions

Inspired by the aforementioned observations, in this paper we propose to modulate the learnable convolution filters via traditional hand-crafted Gabor filters, aiming to reduce the number of network parameters and enhance the robustness of learned features to orientation and scale changes. Steerable and scalable kernels like Gabor functions are advantageous in the sense that they can be specified and computed easily. Specifically, in each convolution layer, the convolution filters are modulated by Gabor filters with different orientations and scales to produce convolutional Gabor orientation filters (GoFs). Compared to the traditional convolution filters, GoFs introduce the additional capability of capturing the visual properties such as spatial localization, orientation selectivity and spatial frequency selectivity in the output feature maps. GoFs are implemented on the basic element of CNNs, *i.e.*, the convolution filter, and thus can be easily integrated into any deep network architecture. DCNNs with GoFs, referred to as GCNs, are able to learn more robust feature representations, particularly for images with spatial transformations. In addition, since GoFs are generated based on a small set of learnable convolution filters, GCNs are more compact and have less parameters than the original CNNs without the need of model compression [10]–[12].

The contributions of this paper are summarized as follows:

- To the best of our knowledge, it is the first attempt to incorporate the Gabor filters into the convolution filter as a modulation process. The modulated filters are able to reinforce the robustness of DCNNs against image transformations such as transitions, scale changes and rotations.
- GoFs can be readily incorporated into different network architectures. GCNs obtain the state-of-the-art performance on various benchmarks when using the conventional CNNs and ResNet [13] as backbone.
- GCNs can significantly reduce the number of learnable filters due to the Gabor filters with pre-defined scales and orientations, making the network more compact while still maintaining high feature representation capacity.
- We provide explicit derivations of updating the Gabor orientation filter in the back-propagation optimization process.

The reminder of this paper is organized as follows. Section 2 reviews related work on Gabor filters and networks designed to enhance the ability to handle scale and rotation variations. Section 3 provides detailed procedures of the proposed Gabor convolution networks by explaining the building blocks and the network optimization. The experimental results of the proposed GCNs on various benchmark datasets for object classification are presented in Section 4. Finally, Section 5 concludes the paper with a few future research directions. The paper is the extension of our conference version [14].

[1]For illustration purpose, AlexNet is selected because its filters are of big sizes.

## II. Related Work

### A. Gabor Filters

Gabor wavelets were invented by Gabor [15] using complex functions to serve as a basis for Fourier transforms in information theory applications. An important property of the wavelets is that the product of its standard deviations is minimized in both time and frequency domains. Gabor filters are widely used to model receptive fields of simple cells of the visual cortex. The Gabor wavelets (kernels or filters) are defined as follows [16]:

$$G(u, v) = \frac{||k_{u,v}||^2}{\sigma^2} e^{-(||k_{u,v}||^2||z||^2/2\sigma^2)}[e^{ik_{u,v}z} - e^{-\sigma^2/2}], \quad (1)$$

where $z = (x, y)$, $k_{u,v} = k_v e^{ik_u}$, $k_v = (\pi/2)/\sqrt{2}^{(v-1)}$, $k_u = u\frac{\pi}{U}$, with $v = 0, \ldots, V$ and $u = 0, \ldots, U$ and $v$ is the frequency and $u$ is the orientation, and $\sigma = 2\pi$.

### B. Learning Feature Representations

Given rich and often redundant convolutional filters, data augmentation is used to achieve local/global transform invariance [17]. Despite the effectiveness of data augmentation, the main drawback is that learning all possible transformations usually requires a large number of network parameters, which significantly increases the training cost and the risk of overfitting. Most recently, TI-Pooling [18] alleviates the drawback by using parallel network architectures for the transformation set and applying the transformation invariant pooling operator on the outputs before the top layer. Nevertheless, with a built-in data augmentation, TI-Pooling requires significantly more training and testing computational cost than a standard CNN.

*1) Spatial Transformer Networks:* To gain more robustness against spatial transformations, a new framework for spatial transformation termed spatial transformer network (STN) [19] is introduced by using an additional network module that can manipulate the feature maps according to the transform matrix estimated with a localization sub-CNN. However, STN does not provide a solution to precisely estimate complex transformation parameters, which are important for some applications such as image registration.

*2) Oriented Response Networks:* By using Actively Rotating Filters (ARFs) to generate orientation-tensor feature maps, Oriented Response Network (ORN) [2] encodes hierarchical orientation responses of discriminative structures. With these responses, ORN can be used to either encode the orientation-invariant feature representation or estimate object orientations. However, ORN is more suitable for small size filters, *i.e.* $3\times3$, whose orientation invariance property is not guaranteed by the ORAlign strategy based on their marginal performance improvement as compared with TI-Pooling.

*3) Deformable Convolutional Network:* Deformable convolution and deformable RoI pooling are introduced in [3] to enhance the transformation modeling capacity of CNNs, making the network robust to geometric transformations. However, the deformable filters also prefer operating on small-sized filters.

*4) Scattering Networks:* In wavelet scattering network [20], [21], expressing receptive fields in CNNs as a weighted sum over a fixed basis allows the new structured receptive field networks to increase the performance considerably over unstructured CNNs for small and medium datasets. In contrast to the scattering networks, our GCNs are based on Gabor filters to change the convolution filters in a steerable way.

*5) Gabor Filters + CNN:* Different pre-processing models such as filters or feature detectors have been employed to improve the accuracy of CNNs. For example, Kwolek [22] proposes a facial regions detection method by combining a Gabor filter and a CNN. Gabor filters are first applied to the face images to extract intrinsic facial features. Then the resulting Gabor feature images are used as input to the CNNs to further extract discriminative features. The similar idea is also explored in [23] for hand-written digit recognition. In essence, the Gabor filter is only used as an offline pre-processing step for the original images. Instead of using the Gabor filter in the data (e.g., images) pre-processing step, there are a few attempts to get rid of the pre-processing overhead by introducing Gabor filters in the convolutional layer of a CNN. For example, in [24], Gabor filters replace the random filter kernels in the 1st convolutional layer. Therefore, only the remaining layers of the CNNs are trained. In this case, the number of learnable parameters are reduced due to the introduction of Gabor filters as the first layer. The same strategy is used in [9], where multiple layers are replaced with Gabor filters to improve the model efficiency.

Compared to [9] and [22]–[24], where Gabor wavelets were used to initialize the deep models or serve as the input layer, we take a different approach by utilizing Gabor filters to modulate the learned convolution filters. And our approach is fundamentally different from the exisiting works that explore Gabor filters with CNNs. Specifically, we change the basic element of CNNs – convolution filters to GoFs to enforce the impact of Gabor filters on each convolutional layer. Therefore, the steerable properties are inherited into the DCNNs to enhance the robustness to scale and orientation variations in feature representations. The filter weights are updated in the proposed Gabor orientation filter in the back-propagation optimization.

## III. Gabor Convolutional Networks

Gabor Convolutional Networks (GCNs) are deep convolutional neural networks using Gabor orientation filters (GoFs). A GoF is a steerable filter, created by manipulating the learned convolution filters via Gabor filter banks, to produce the enhanced feature maps. With GoFs, GCNs not only have significant fewer filter parameters to learn, but also lead to enhanced deep models.

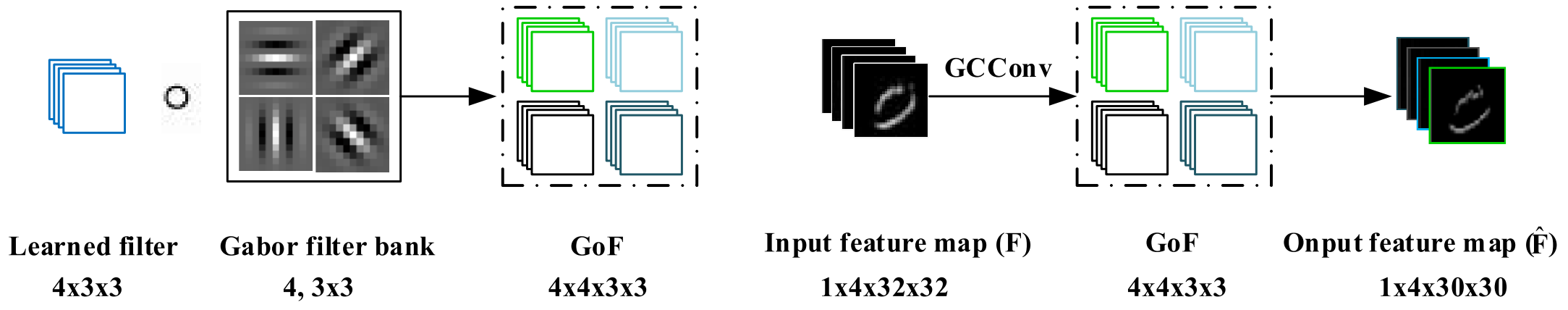

Fig. 2. Left shows modulation process of GoFs. Right illustrates an example of GCN convolution with 4 channels. In a GoF, the number of channels is set to be the number of Gabor orientations $U$ for implementation convenience.

In what follows, we address three issues in implementing GoFs in DCNNs. First, we give the details on obtaining GoFs through Gabor filters. Second, we describe convolutions that use GoFs to produce feature maps with scale and orientation information enhanced. Third, we show how GoFs are learned during the back-propagation update stage.

### A. Convolutional Gabor Orientation Filters (GoFs)

Gabor filters are of $U$ directions and $V$ scales. To incorporate the steerable properties into the GCNs, the orientation information is encoded in the learned filters, and at the same time the scale information is embedded into different layers. Due to the orientation and scale information captured by Gabor filters in GoFs, the corresponding convolution features are enhanced.

Before being modulated by Gabor filters, the convolution filters in standard CNNs are learned by back propagation (BP) algorithm, which are denoted as learned filters. Unlike standard CNNs, to encode the orientation channel, the learned filters in GCNs are three-dimensional. Let a learned filter be of size $N \times W \times W$, where $W \times W$ is the size of the filter and $N$ refers to channel. If the dimensions of the weight per layer in traditional CNNs is expressed as $C_{out} \times C_{in} \times W \times W$, GCNs will represent it as $C_{out} \times C_{in} \times N \times W \times W$, $C_{out}$ and $C_{in}$ represent the channel of output and input feature map respectively. To keep the channels quantity of the feature map consistent during the forward convolution process, $N$ is chosen to be $U$, which is the number of orientations of the Gabor filters that will be used to modulate this learned filter. A GoF is obtained based on a modulated process using $U$ Gabor filters on the learned filters for a given scale $v$. The details concerning the filter modulation are shown in Eq. 2 and Fig. 2. For the $v$th scale, we define:

$$C_{i,u}^{v} = C_{i,o} \circ G(u,v), \tag{2}$$

where $C_{i,o}$ is a learned filter. $G(u,v)$ represents a group of Gabor filters with different orientations and scales, which is defined in Eq.1, and $\circ$ is an element-by-element product operation between $G(u,v)$[2] and each 2D filter of $C_{i,o}$. $C_{i,u}^{v}$ is the modulated filter of $C_{i,o}$ by the $v$-scale Gabor filter $G(u,v)$. Then a GoF is defined as:

$$C_{i}^{v} = (C_{i,1}^{v}, \ldots, C_{i,U}^{v}). \tag{3}$$

[2]The real parts of Gabor filters are used.

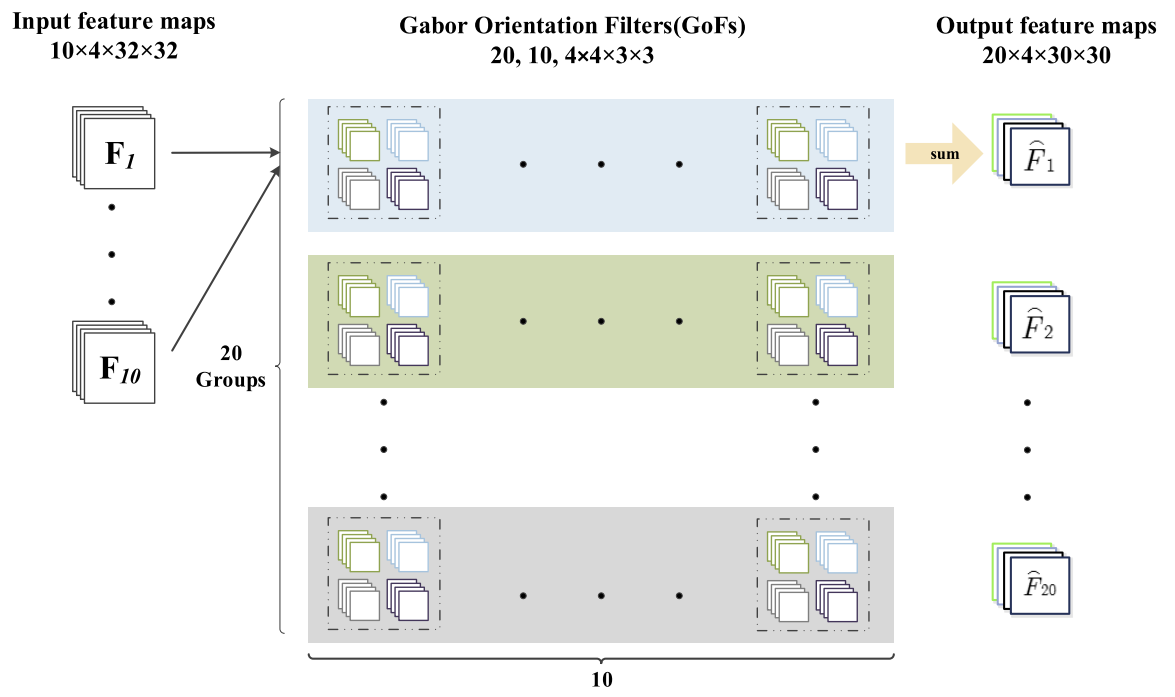

Fig. 3. Forward convolution process of GCNs with multiple feature maps. There are 10 and 20 feature maps in the input and the output, respectively. The reconstructed filters are divided into 20 groups and each group contains 10 reconstructed filters.

Thus, the $i$th GoF $C_i^v$ is actually $U$ 3D filters (see Fig. 2, where $U = 4$). In GoFs, the value of $v$ increases with increasing layers, which means that scales of Gabor filters in GoFs are changed based on layers. At each scale, the size of a GoF is $U \times N \times W \times W$. However, we only save $N \times W \times W$ learned filters because the Gabor filters are given, which means that we can obtain enhanced features by this modulation without increasing the number of parameters. To simplify the description of the learning process, $v$ is omitted in the next section.

### B. GCN Convolution

In GCNs, GoFs are used to produce feature maps, which explicitly enhance the scale and orientation information in deep features. A output feature map $\widehat{F}$ in GCNs is denoted as:

$$\widehat{F} = GCconv(F, C_i), \tag{4}$$

where $C_i$ is the $i$th GoF and $F$ is the input feature map as shown in Fig. 2. The channels of $\widehat{F}$ are obtained by the following convolution:

$$\widehat{F}_{i,k} = \sum_{n=1}^{N} F^{(n)} \otimes C_{i,u=k}^{(n)}, \tag{5}$$

where $(n)$ refers to the $n$th channel of $F$ and $C_{i,u}$, and $\widehat{F}_{i,k}$ is the $k$th orientation response of $\widehat{F}_i$. For example as shown in Fig. 2, let the size of the input feature map be $1 \times 4 \times 32 \times 32$. If there are 10 GoFs with 4 Gabor orientations, the size of the output feature map is $10 \times 4 \times 30 \times 30$. Figure 3 shows the forward convolution process of GCNs when the input feature map extended to the multiple channel ($C_{in} \neq 1$). A visualization example of the first convolution layer in GCNs, which is used in Section IV, is presented in Fig. 4.

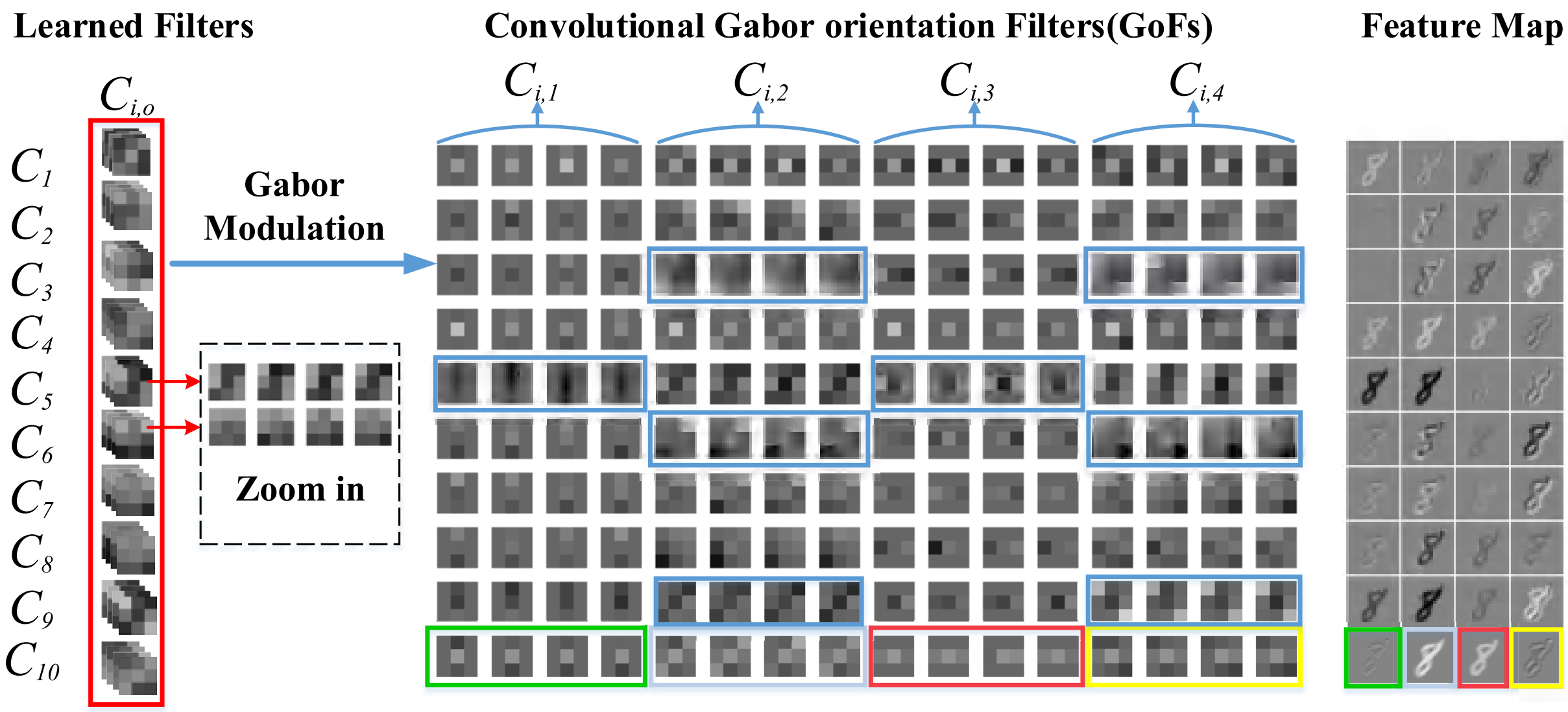

Fig. 4. Visualization of the first convolution layer of GCNs. Each row represents a group of GoFs and its corresponding feature map. *i.e.* the 10th GoFs ($C_{10,1} \ldots C_{10,4}$). Each 4-orientation channel GoF is labeled as different colors. The output feature map also has 4-orientation channel, which is labeled with the same color as its corresponding GoF. The examples in the blue rectangle show that GOFs carry various orientation information.

### *C. Updating GoF*

Different from traditional CNNs, the weights involved in the forward calculation are GOFs in GCNs, but the weights which are saved are only the learned filters. Therefore, in the back-propagation (BP) process, only the leaned filer $C_{i,o}$ needs to be updated. We need to sum up the gradient of the sub-filters in GOFs to the corresponding learned filters, and we have:

$$\delta = \frac{\partial L}{\partial C_{i,o}} = \sum_{u=1}^{U} \frac{\partial L}{\partial C_{i,u}} \circ G(u,v) \tag{6}$$

$$C_{i,o} = C_{i,o} - \eta\delta, \tag{7}$$

where $L$ is the loss function. From the above equations, it can be seen that the BP process is easily implemented and is very different from ORNs and deformable kernels that usually require a relatively complicated procedure. By only updating the learned convolution filters $C_{i,o}$, the GCNs model is more compact and efficient, and also is more robust to orientation and scale variations. The pseudo code of GCNs is shown in Algorithm 1.

## IV. Implementation and Experiments

In this section, the details of the GCNs implementation are elaborated based on conventional CNNs, and ResNet architectures as shown in Fig. 5. We evaluate our GCNs (CNNs) on the MNIST dataset [25], [26], and also its rotated version MNIST-rot dataset generated by rotating each sample in MNIST by a random angle between $[0, 2\pi]$. We further validate the effectiveness of GCNs (ResNet) on the SVHN dataset [27], CIFAR-10 and CIFAR-100 [28], ImageNet2012 [29], and Food-101 [30]. In our experiments, We have two GPU platforms used in our experiments, NVIDIA GeForce GTX 1070 and GeForce GTX TITAN X(2).

**Algorithm 1** Gabor Convolutional Networks

Initializing the hyper-parameters and network structure parameters, whose details can also refer to our source code.
Set $U$ and $V$ ($U$ and $V$ refers to the number of Gabor orientations and scales respectively.)
Start training:
**repeat**
    Inputting a mini-batch set of training images.
    Producing GOFs by learned filters and Gabor filters using Eq. 3.
    GCNs forward convolution based on the input feature maps and GoFs using Eq. 5. And the fully connected features are further obtained.
    Calculating the cross-entropy loss and then performing back propagation process according to Eq. 6
    Updating learned filters based on Eq. 7.
**until** the maximum epoch.

### *A. MNIST*

For the MNIST dataset, we randomly select 10,000 samples from the training set for validation and the remaining 50,000 samples for training. Adadelta optimization algorithm [31] is used during the training process, with the batch size as 128, initial learning rate as 0.001 ($\eta$). The learning weight decay is set as 0.00005, and the learning rate is reduced to half per 25 epochs. The state-of-the-art STN [19], TI-Pooling [18], ResNet [13] and ORNs [2] are exploited for comparison. Among them, STN is more robust to

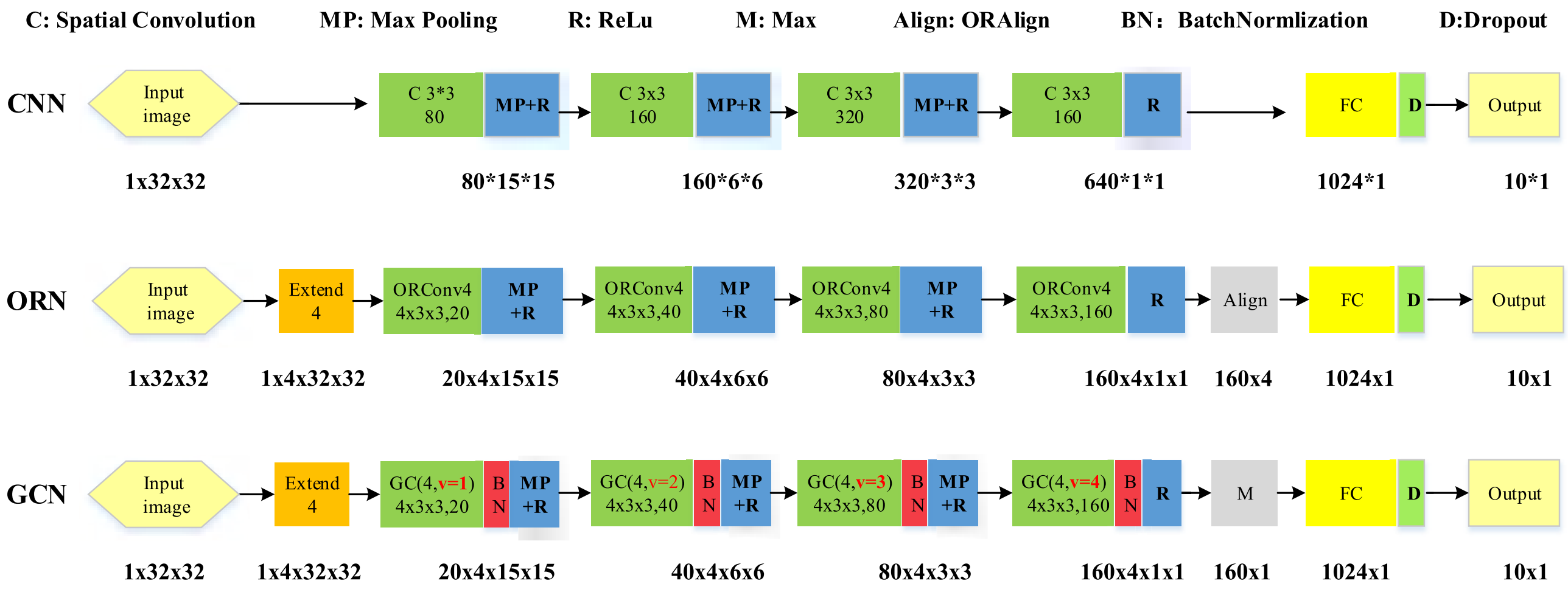

Fig. 5. Network structures of CNNs, ORNs and GCNs.

TABLE I

RESULTS (ERROR RATE (%) ON MNIST) VS. GABOR FILTER SCALES. THE MORE SCALES ($V = 4$) IN USE CAN LEAD TO BETTER PERFORMANCE THAN THE SINGLE SCALE ($V = 1$) UPON DIFFERENT CONVOLUTION FILTER SIZES

| kernel size | 5x5 | 3x3 |
|---|---|---|
| # params (M) | 1.86 | 0.78 |
| $V = 1$ scale | 0.48 | 0.51 |
| $V = 4$ scales | 0.48 | 0.49 |

TABLE II

RESULTS (ERROR RATE (%) ON MNIST) VS. GABOR FILTER ORIENTATIONS

| $U$ | 2 | 3 | 4 | 5 | 6 | 7 |
|---|---|---|---|---|---|---|
| 5x5 | 0.52 | 0.51 | 0.48 | 0.49 | 0.49 | 0.52 |
| 3x3 | 0.68 | 0.58 | 0.56 | 0.59 | 0.56 | 0.6 |

spatial transformation than the baseline CNNs, due to a spatial transform layer prior to the first convolution layer. TI-Pooling generates training samples in 8 directions by data argumentation, thus having 8 networks and each corresponding to one direction. And also TI-Pooling adopts a transform-invariant pooling layer to get the response of main direction, resulting in rotation robust features. ORNs capture the response of each direction by rotating the original convolution kernel spatially. Different from conventional CNNs, both GCNs and ORNs extend the 2D convolution kernel to the 3D kernel. STN, TI-Pooling and ORNs are the-state-of-art deep learning algorithms to recognize the rotated digits.

Fig. 5 illustrates the network architectures of CNNs, ORNs and GCNs ($U = 4$) used in this experiment. To be fair for above models, we adopt Max-pooling and ReLU after convolution layers, and a dropout layer [32] after the fully connected (FC) layer to avoid over-fitting. To compare with other CNNs in a similar model size, we reduce the width of layer[3] by a certain proportion as done in ORNs, *i.e.* 1/8 [2].

We evaluate different scales for different GCNs layers (*i.e.* $V = 4$, $V = 1$), where larger scale Gabor filters are used in shallow layers or a single scale is used in all layers. It should be noted that in the following experiments, we also use $V = 4$ for deeper networks (ResNet). As shown in Table I, the results of $V = 4$ in terms of error rate are better than those when a single scale ($V = 1$) is used in all layers. We also test different orientations as shown in Table II. The results indicate that GCNs perform better using 3 to 6 orientations when $V = 4$, which is more flexible than ORNs. In comparison, ORNs use a complicated interpolation process via ARFs besides 4- and 8-pixel rotations. GCNs achieve different results when $U$ is set from 2 to 7, which show that the number of orientation channel should not be too large or too small. Too few orientation channels are not able to extract enough informative features, while too many orientation channels may make the network too complex. When $U = 4, 5$, the results are more stable.

In Table III, the second column refers to the width of each layer, and a similar notation is also used in [33]. Considering a GoF has multiple channels ($N$), we decrease the width of layer (*i.e.* the number of GoFs per layer) to reduce the model size to facilitate a fair comparison. The parameter size of GCNs is linear with channel ($N$) but quadratic with width of layer. Therefore, the GCNs complexity is reduced as compared with CNNs (see the third column of Table III). In the fourth column, we compare the computation time (s) for training epoch of different methods using GTX 1070, which clearly shows that GCNs are more efficient than other state-of-the-art models. The performance comparison is shown in the last two columns in terms of error rate. By comparing with baseline CNNs, GCNs achieved much better performance with 3x3 kernel but only using 1/12, 1/4 parameters of CNNs. It is observed from the experiments that GCNs with 5x5 and 7x7 kernels achieve test errors of 1.10% on MNIST-rot and 0.42% on MNIST, respectively, which are better than those of ORNs. This can be explained by the fact that the kernels with larger size carry more information of Gabor orientation, and thus capture better orientation response features. Table III also demonstrates that a larger GCNs model can result in

[3]The number of convolution kernels per layer.

TABLE III

RESULTS COMPARISON ON MNIST

| Method | | | | error (%) | |
|---|---|---|---|---|---|
| | # network stage kernels | # params (M) | time (s) | MNIST | MNIST-rot |
| Baseline CNN | 80-160-320-640 | 3.08 | 6.50 | 0.73 | 2.82 |
| STN | 80-160-320-640 | 3.20 | 7.33 | 0.61 | 2.52 |
| TIPooling$_{(\times 8)}$ | (80-160-320-640)$_{\times 8}$ | 24.64 | 50.21 | 0.97 | not permitted |
| ORN4(ORAlign) | 10-20-40-80 | 0.49 | 9.21 | 0.57 | 1.69 |
| ORN8(ORAlign) | 10-20-40-80 | 0.96 | 16.01 | 0.59 | 1.42 |
| ORN4(ORPooling) | 10-20-40-80 | 0.25 | 4.60 | 0.59 | 1.84 |
| ORN8(ORPooling) | 10-20-40-80 | 0.39 | 6.56 | 0.66 | 1.37 |
| GCN4(with $3 \times 3$) | 10-20-40-80 | 0.25 | 3.45 | 0.63 | 1.45 |
| GCN4(with $3 \times 3$) | 20-40-80-160 | 0.78 | 6.67 | 0.56 | 1.28 |
| GCN4(with $5 \times 5$) | 10-20-40-80 | 0.51 | 10.45 | 0.49 | 1.26 |
| GCN4(with $5 \times 5$) | 20-40-80-160 | 1.86 | 23.85 | 0.48 | **1.10** |
| GCN4(with $7 \times 7$) | 10-20-40-80 | 0.92 | 10.80 | 0.46 | 1.33 |
| GCN4(with $7 \times 7$) | 20-40-80-160 | 3.17 | 25.17 | **0.42** | 1.20 |

TABLE IV

RESULTS COMPARISON ON SVHN. NO ADDITIONAL TRAINING SET IS USED FOR TRAINING

| Method | VGG | ResNet-110 | ResNet-172 | GCN4-40 | GCN4-28 | ORN4-40 | ORN4-28 | GCN4-40-Gau |
|---|---|---|---|---|---|---|---|---|
| # params | 20.3M | 1.7M | 2.7M | 2.2M | 1.4M | 2.2M | 1.4M | 2.2M |
| Accuracy (%) | 95.66 | 95.8 | 95.88 | 96.9 | 96.86 | 96.35 | 96.19 | 96.1 |

better performance. In addition, on the MNIST-rot datasets, the performance of baseline CNNs model is greatly affected by rotation, while ORNs and GCNs can capture orientation features and achieve better results. Again, GCNs outperform ORNs, which confirms that Gabor modulation indeed helps to gain the robustness to rotation variations. This improvement is attributed to the enhanced deep feature representations of GCNs based on the steerable filters. In contrast, ORNs only actively rotate the filters and lack a feature enhancement process.

### B. SVHN

The Street View House Numbers (SVHN) dataset [27] is a real-world image dataset taken from Google Street View images. SVHN contains MNIST-like 32x32 images centered around a single character, which however include a plethora of challenges like illumination changes, rotations and complex backgrounds. The dataset consists of 600000 digit images: 73257 digits for training, 26032 digits for testing, and 531131 additional images. Note that the additional images are not used for all methods in this experiment. For this large scale dataset, we implement GCNs based on ResNet. Specifically, we replace the spatial convolution layers with our GoFs based GCConv layers, leading to GCN-ResNet. The bottleneck structure is not used since the 1x1 kernel does not propagate any Gabor filter information. ResNet divides the whole network into 4 stages, and the width of stage (the number of convolution kernels per layer) is set as 16, 16, 32, and 64, respectively. We make appropriate adjustments to the network depth and width to ensure our GCNs method has a similar model size as compared with VGG [34] and ResNet. We set up 40-layer and 28-layer GCN-ResNets with basic block-(c)(Fig. 6), using the same hyper-parameters as ResNet. The network stage is also set as 16-16-32-64. The results are listed in Table IV. Compared to VGG model, GCNs have much smaller parameter size, yet obtain a better performance with 1.2% improvement. With a similar parameter size, the GCN-ResNet achieves better results (1.1%, 0.66%) than ResNet and ORNs respectively, which further validates the superiority of GCNs for real-world problems. We also compare with GCN4-Gau, where Gabor filter banks is replaced by the Gaussian mask that is same as in Gabor filters. It shows that the GCNs can still achieve better performances than those of GCN4-Gau, which confirmed that Gabor filters can really improve the performance of the CNNs.

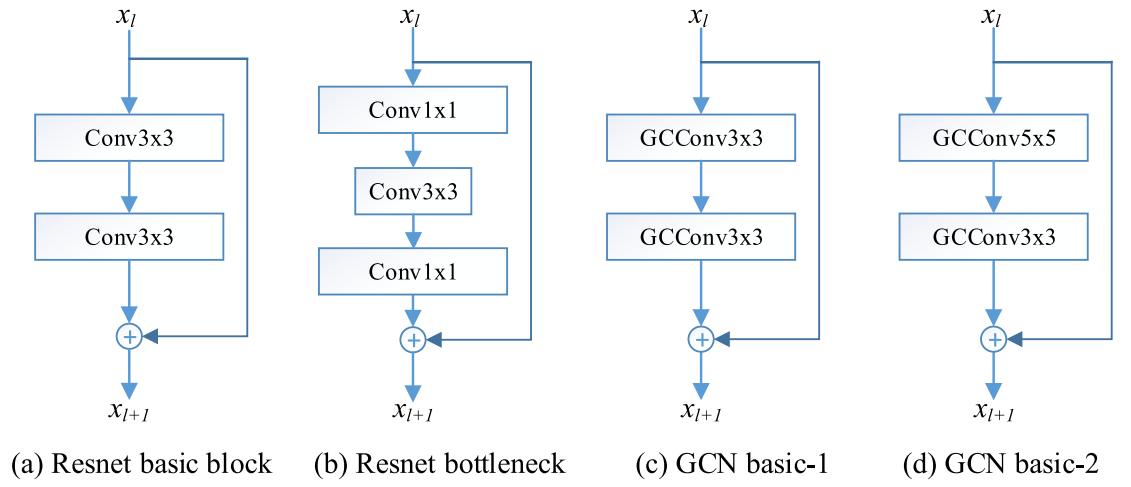

Fig. 6. The residual block. (a) and (b) are for ResNet. (c) Small kernel and (d) large kernel are for GCNs.

### C. Natural Image Classification

For the natural image classification task, we use the CIFAR datasets including CIFAR-10 and CIFAR-100 [28]. The CIFAR datasets consist of 60000 color images of size 32x32 in 10 or 100 classes, with 6000 or 600 images per class. There are 50000 training images and 10000 test images.

CIFAR datasets contain a wide variety of categories with object scale and orientation variations. Similar to SVHN, we test GCN-ResNet on CIFAR datasets. Experiments are conducted to compare our method with the state-of-the-art networks (*i.e.* NIN [35], VGG [34], ORN [2] and ResNet [13]). On CIFAR-10, Table V shows that GCNs consistently improve the performance regardless of the number of parameters or kernels as compared with the baseline ResNet. We further compare GCNs with the Wide Residue network (WRN) [33], and again GCNs achieve a better result (3.88% vs. 4% error rate) when our model is half the size of WRN, indicating significant advantage of GCNs in terms of model efficiency. Similar to CIFAR-10, one can also observe the performance improvement on CIFAR-100, with similar parameter sizes. Moreover, when using different kernel size configurations (from $3 \times 3$ to $5 \times 5$ as shown in Fig. 6(c) and Fig. 6(d)), the model size is increased but with a performance (error rate) improvement from 6.34% to 5.62%. We notice that some top improved classes in CIFAR10 are bird (4.1% higher than the baseline ResNet), and deer (3.2%), which exhibit significant within class scale variations. This implies that the Gabor filter modulation in CNNs enhances the capability of handling scale variations (see Fig. 7). Training loss and test error curves are shown in Fig. 8.

TABLE V
RESULTS COMPARISON ON CIFAR-10 AND CIFAR-100

| Method | | | error (%) CIFAR-10 | CIFAR-100 |
|---|---|---|---|---|
| NIN | | | 8.81 | 35.67 |
| VGG | | | 6.32 | 28.49 |
| | # network stage kernels | # params Fig.6(c)/Fig.6(d) | | |
| ResNet-110 | 16-16-32-64 | 1.7M | 6.43 | 25.16 |
| ResNet-1202 | 16-16-32-64 | 10.2M | 7.83 | 27.82 |
| GCN2-110 | 12-12-24-45 | 1.7M/3.3M | 6.34/5.62 | |
| GCN2-110 | 16-16-32-64 | 3.4M/6.5M | 5.65/4.96 | 26.14/25.3 |
| GCN4-110 | 8-8-16-32 | 1.7M | 6.19 | |
| GCN2-40 | 16-32-64-128 | 4.5M | 4.95 | 24.23 |
| GCN4-40 | 16-16-32-64 | 2.2M | 5.34 | 25.65 |
| WRN-40 | 64-64-128-256 | 8.9M | 4.53 | 21.18 |
| WRN-28 | 160-160-320-640 | 36.5M | 4.00 | **19.25** |
| GCN2-40 | 16-64-128-256 | 17.9M | 4.41 | 20.73 |
| GCN4-40 | 16-32-64-128 | 8.9M | 4.65 | 21.75 |
| GCN3-28 | 64-64-128-256 | 17.6M | **3.88** | 20.13 |

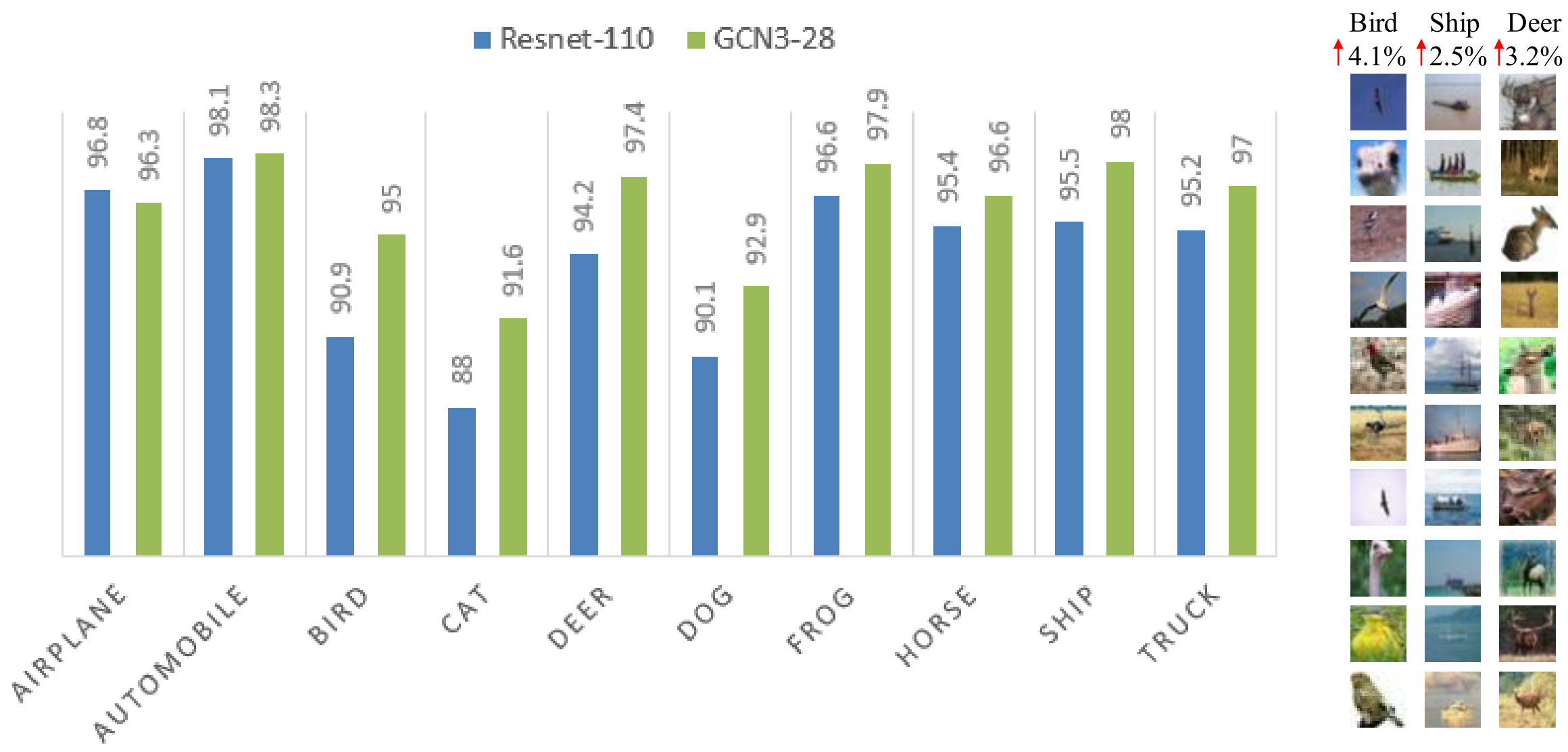

Fig. 7. Recognition results of different categories on CIFAR10. Compared with ResNet-110, GCNs perform significantly better on the categories with large scale variations.

### D. Large Size Image Classification

The previous experiments are conducted on datasets with small size images (*e.g.,* $32 \times 32$ for the SVHN dataset). To further show the effectiveness of the proposed GCNs method, we evaluate it on the ImageNet [29] dataset. Different from MNIST, SVHN and CIFAR, ImageNet consists of images with a much higher resolution. In addition, the images usually contain more than one attribute per image, which may have a large impact on the classification accuracy. We first choose a 100-class ImageNet2012 [29] subset in this experiment. The 100 classes are selected from the full ImageNet dataset at a step of 10. This subset is also applied in [36], [37].

For the ImageNet-100 experiment, we train a 34-layer GCN with 4-orientation channels, and the scale setting is the same as previous experiments. A ResNet-101 model is set as the baseline. Both GCNs and ResNet are trained after 120 epochs. The learning rate is initialized as 0.1 and decreases to 1/10 times per 30 epochs. Top-1 and Top-5 errors are used as evaluation metrics. The convergence of ResNet is better at the early stage. However, it tends to be saturated quickly. GCNs have a slightly slower convergence speed, but show better performance in later epochs. The test error curve is depicted in Fig. 9. As compared to the baseline, our GCNs achieve better classification performances (*i.e.*, Top-5 error: 3.04% vs. 3.16%, Top-1 error: 11.46% vs. 11.94%) when using fewer parameters (35.85M vs. 44.54M). We also train a 34-layer model based on full ImageNet. Compared with ResNet-34 (71.6%), we obtain a better performance (73.2%), which further validate the effectiveness of our method.

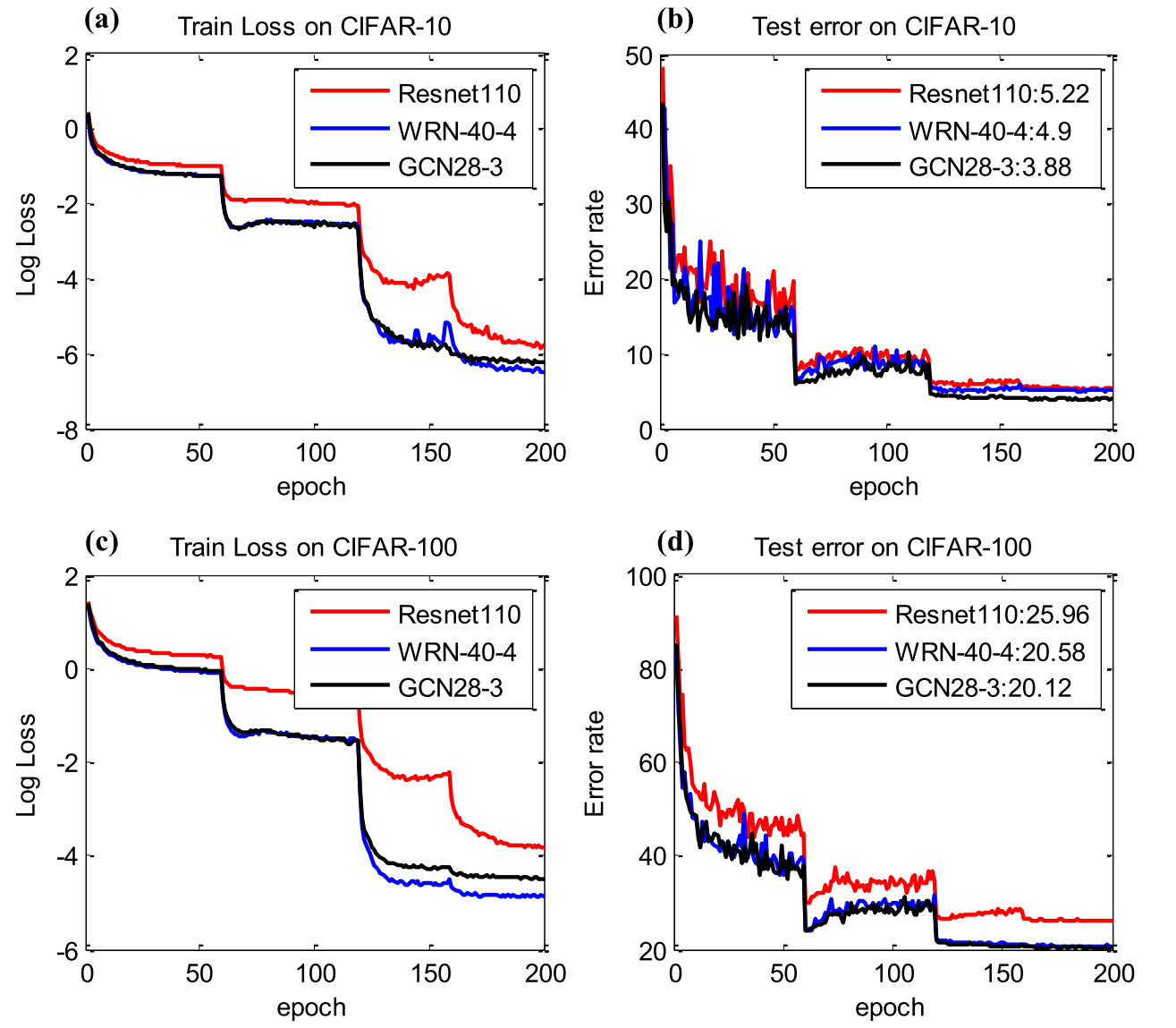

Fig. 8. Training loss and test error curves on CIFAR dataset. (a),(b) for CIFAR-10, (c),(d) for CIFAR-100. Compared with baseline ResNet-110, GCN achieved a faster convergence speed and lower test error. WRN and GCN achieved similar performence, but GCN had lower error rate on the test set.

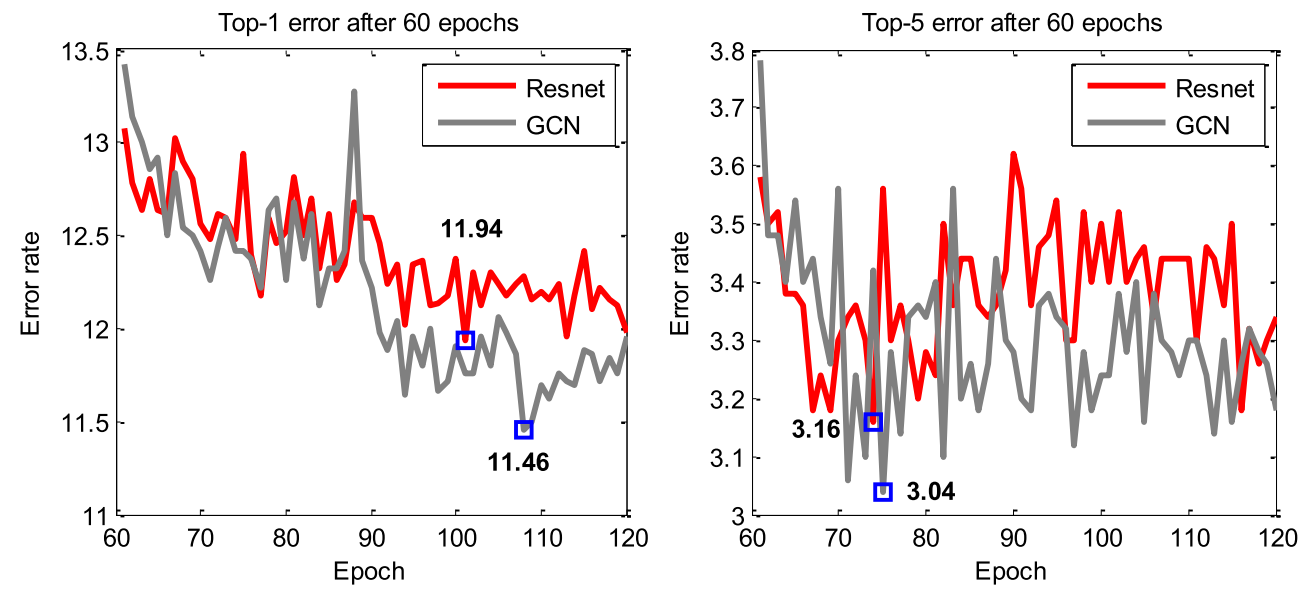

Fig. 9. Test error curve for the ImageNet experiment.The first 60 epochs are omitted for clarity.

### *E. Experiment on Food-101 Dataset*

The Food-101 dataset was respectively collected for food recognition based on images with much higher resolutions, which still challenge existing methods. We further validate the effectiveness of our TaylorNets on both of them, by comparing with state-of-the-art ResNets and the kernel pooling methods [38]. It should be noted that the kernel pooling method is also based on detailed information.

TABLE VI
COMPARISON (ERROR (%)) WITH STATE-OF-ART NETWORKS ON FOOD-101

| | ResNet-50 | CBP [39] | KP [38] | GCNs |
|---|---|---|---|---|
| Food-101 | 17.9 | 16.8 | 14.5 | 14.2 |

We train a 28-layer GCNs, and the network stage is set to 16-16-32-64. The comparative results are listed in Table VI. It is observed that, our GCNs achieves a much better performance than the state-of-the-art kernel pooling method in te rms of error rate. The reason might liein that our method c an capture nonlinear features based onsteerable filters, whic h benefits the food recognition sufferingfrom the noise probl ems.

## V. CONCLUSION

This paper has presented a new end-to-end deep model by incorporating Gabor filters to DCNNs, aiming to enhance the deep feature representations with steerable orientation and scale capacities. The proposed Gabor Convolutional Networks (GCNs) improve DCNNs on the generalization ability of rotation and scale variations by introducing extra functional modules on the basic element of DCNNs, *i.e.*, the convolution filters. GCNs can be easily implemented using popular architectures. The extensive experiments show that GCNs significantly improved baselines, resulting in the state-of-the-art performance over several benchmarks. In the future, more GCNs architectures (larger ones) will be tested on other tasks, such as object tracking, detection and segmentation [40]–[43].

**Shangzhen Luan** received the B.S. and master's degrees from the Department of Automation Science and Electrical Engineering, Beihang University. His current research interests include signal and image processing, pattern recognition, and computer vision.

**Chen Chen** received the B.E. degree in automation from Beijing Forestry University, Beijing, China, in 2009, the M.S. degree in electrical engineering from Mississippi State University, Starkville, MS, USA, in 2012, and the Ph.D. degree from The University of Texas at Dallas, Richardson, TX, USA, in 2016. He is currently a Post-Doctoral Fellow with the Center for Research in Computer Vision, University of Central Florida, Orlando, FL, USA. His current research interests include compressed sensing, signal and image processing, pattern recognition, and computer vision. He has authored or co-authored over 40 papers in refereed journals and conferences in these areas.

**Baochang Zhang** received the B.S., M.S., and Ph.D. degrees in computer science from the Harbin Institute of the Technology, Harbin, China, in 1999, 2001, and 2006, respectively. From 2006 to 2008, he was a Research Fellow with The Chinese University of Hong Kong, Hong Kong, and also with Griffith University, Brisban, Australia. He is currently an Associate Professor with Beihang University, Beijing, China. His current research interests include pattern recognition, machine learning, face recognition, and wavelets.

**Jungong Han** was a Senior Scientist with Civolution Technology (a combining synergy of Philips CI and Thomson STS) from 2012 to 2015, a Research Staff with the Centre for Mathematics and Computer Science from 2010 to 2012, and a Researcher with the Technical University of Eindhoven, The Netherlands, from 2005 to 2010. He is currently a Senior Lecturer with the Department of Computer Science, Northumbria University, U.K.

**Jianzhuang Liu** received the Ph.D. degree in computer vision from The Chinese University of Hong Kong, Hong Kong, in 1997. From 1998 to 2000, he was a Research Fellow with Nanyang Technological University, Singapore. From 2000 to 2012, he was a Post-Doctoral Fellow, an Assistant Professor, and an Adjunct Associate Professor with The Chinese University of Hong Kong. In 2011, he joined the Shenzhen Institutes of Advanced Technology, Chinese Academy of Sciences, Shenzhen, China, as a Professor. He is currently a Principal Researcher with Huawei Technologies Company, Ltd., Shenzhen. He has authored over 150 papers. His research interests include computer vision, image processing, machine learning, multimedia, and graphics.